# 🔥FLamE: Few-shot Learning from Natural Language Explanations

**Yangqiaoyu Zhou**  **Yiming Zhang**  **Chenhao Tan**

University of Chicago

`{zhouy1, yimingz0, chenhao}@uchicago.edu`

## Abstract

Natural language explanations have the potential to provide rich information that in principle guides model reasoning. Yet, recent work by Lampinen et al. (2022) has shown limited utility of natural language explanations in improving classification. To effectively learn from explanations, we present **FLamE**, a two-stage few-shot learning framework that first generates explanations using GPT-3, and then fine-tunes a smaller model (e.g., RoBERTa) with generated explanations. Our experiments on natural language inference demonstrate effectiveness over strong baselines, increasing accuracy by 17.6% over GPT-3 Babbage and 5.7% over GPT-3 Davinci in e-SNLI. Despite improving classification performance, human evaluation surprisingly reveals that the majority of generated explanations does not adequately justify classification decisions. Additional analyses point to the important role of label-specific cues (e.g., "`not know`" for the neutral label) in generated explanations.

## 1 Introduction

Collecting and learning from natural language explanations has received increasing attention in the NLP community (Wiegreffe and Marasović, 2021). The idea of learning from natural language explanations is especially appealing in few-shot learning because explanations can provide rich information about the task and guide model reasoning when there are limited supervision signals.

Although large-scale language models (LLMs) have demonstrated a remarkable capability in few-shot learning (Brown et al., 2020; Rae et al., 2022; Chowdhery et al., 2022a), the effect of learning from natural language explanations remains mixed. On the one hand, Wei et al. (2022b) demonstrates impressive success with chain-of-thought prompting, especially in arithmetic reasoning. On the other hand, in a systematic evaluation of the effect of explanations on in-context learning, Lampinen et al. (2022) discover only a marginal improvement from explanations, even when experimenting with massive models (280B). It thus remains an open question how we can leverage LLMs to effectively learn from natural language explanations.

We propose a two-stage framework (🔥**FLamE**) for <u>F</u>ew-shot <u>L</u>earning fro<u>m</u> natural language <u>E</u>xplanations. Fig. 1 gives a graphical overview of our approach. First, our framework leverages the ability of large-scale language models (e.g., GPT-3) to generate explanations. Second, it uses explanation-aware prompt-based classification where we can fine-tune a smaller model (e.g., RoBERTa). The second step enables the model to tailor to the imperfect explanations from GPT-3 and also opens up opportunities to interpret and probe the model given its transparent internals.

We show that **FLamE** outperforms strong baselines in natural language inference. Compared to GPT-3 finetuned with explanations, **FLamE** achieves higher accuracy than Babbage by 17.6% on e-SNLI and 6.9% on e-HANS, and also outperforms Davinci by 14.2% on e-SNLI and 14.3% on e-HANS. In addition, **FLamE** outperforms the strongest baselines that do not use explanations by 5.7% on e-SNLI and 1.2% on e-HANS.

Furthermore, we conduct an in-depth analysis to understand how our approach improves classification and reveal the important role of label-specific cues. We first show that the generated explanations do not perform valid inferences according to human evaluation. This result corroborates recent work on the characteristics of GPT-3 explanations: they read fluent but lack accurate reasoning (Wiegreffe et al., 2022; Ye and Durrett, 2022). We also observe that GPT-3 explanations frequently include tokens that encode label information (e.g., "`not know`" for the neutral label).

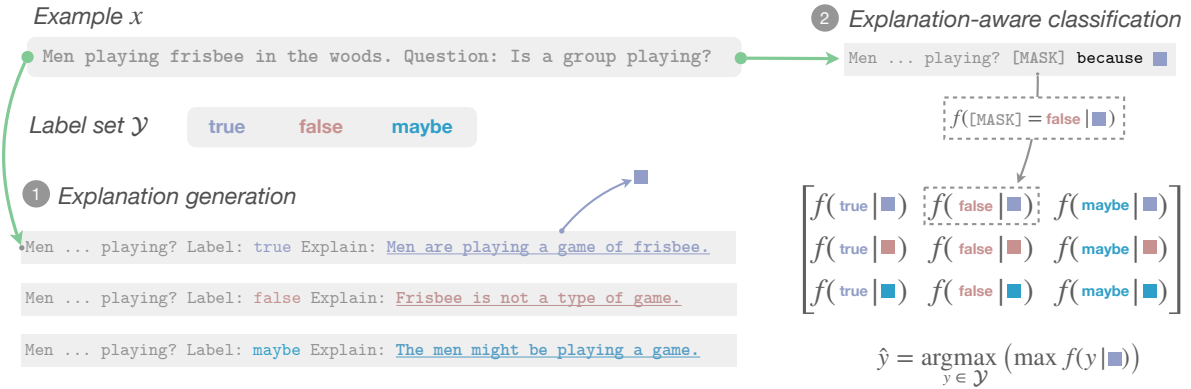

Figure 1: An example illustrating the two-stages of **FLamE**: (1) explanation generation and (2) explanation-aware classification. We use distinct colors to represent labels, and use ■ to indicate a generated explanation. In stage 1, **FLamE** generates an explanation for each label y ∈ 𝒴 with GPT-3. In stage 2, **FLamE** uses a prompt-based model to classify with the aid of explanations. Specifically, for each label y and generated explanation ■, we measure the (unnormalized) probability f(y|■) of unmasking y from the prompt in the presence of ■, and the predicted label ŷ is the label associated with the maximum probability in the matrix.

Our two-staged framework uses a small classification model, enabling us to probe the behavior of our model with perturbed explanations. To investigate the reliance of our model on label-specific cues, we perturb explanations during test time (by changing nouns and verbs), to remove relevant information for the task while keeping label cues. Although these perturbed explanations are *not* related to the original premise and hypothesis, we find that our classification model still makes the same prediction. This observation confirms that generating label-specific cues is the key reason that imperfect explanations manage to improve classification performance.

It is worth noting that our main experiments were done with the GPT-3's fine-tuning API due to our preliminary experiments and budget considerations. We later found that our performance improvement in e-SNLI is robust against GPT-3 in-context learning with Davinci and Babbage, but it is not against GPT-3 Davinci in e-HANS, likely due to the templated nature of e-HANS. This discrepancy between in-context learning and fine-tuning with GPT-3 motivates future work to understand and control these black-box models.

In summary, our contributions are:

- We propose **FLamE**, a few-shot learning framework that effectively leverages natural language explanations to improve classification.

- Our analysis reveals the limitations of generated explanations and sheds light on how illogical explanations could help.

- Our framework enables probing experiments to understand the behavior of a classification pipeline with large-scale language models.

## 2 Learning from Explanations

Our method (**FLamE**) consists of two stages: 1) *explanation generation* with GPT-3 and 2) *explanation-aware classification* with a smaller standalone model (Fig. 1). Deviating from the paradigm in literature of treating both processes as a joint optimization problem (Hase et al., 2020), the disentanglement of explanation generation from classification allows our methods to use the capability of large language models to generate fluent explanations from a handful of examples, while leaving classification to a downstream model, thereby enabling probing experiments and explicit control over the classification component.

### 2.1 Explanation Generation

A key issue with training a few-shot model with the gold explanations as input is that explanations are unlikely to be available at test time. Training with gold explanations and testing in its absence leads to a distribution shift between training and inference. To make explanations available at test time, **FLamE** uses GPT-3 for explanation generation.

Following prior work (Camburu et al., 2018; Wei et al., 2022b), we consider two ways of generating explanations with GPT-3. One approach is to simply prompt GPT-3 models with a test instance without label information.[1] We experiment with this mode of explanation generation, dubbed *explain-then-predict* following Camburu et al. (2018).

As a valid explanation must explain the correct classification decision, trying to generate an explanation without the correct label essentially shifts the burden of classification to the explainer. Indeed, we observe that even GPT-3 Davinci struggles to produce reasonable explanations when the correct label is not given. Similar to our observation, Wiegreffe et al. (2020) find labels are necessary for generating high-quality explanations.

To address the dependency of explanation generation on the ground truth, we use an additional generation scheme, *predict-then-explain*, in which we generate an explanation $\hat{e}_y$ targeting every label $y \in \mathcal{Y}$. In Fig. 1(1), we provide an example illustrating the *predict-then-explain* scheme.[2]

## 2.2 Classification with Explanations

Our few-shot classification framework extends pattern-exploiting training (PET), a performant few-shot classification framework proposed by Schick and Schütze (2020). The key intuition is to convert a classification problem into a slot-filling problem to leverage the knowledge encoded in pretrained language models. We refer the interested reader to Appendix A for an overview of the PET framework.

To incorporate explanations into the PET framework, we propose *explanation-aware patterns* $\text{EP}: \mathcal{X} \times \mathcal{E} \rightarrow \mathcal{V}^\star$. $\text{EP}$ converts an example x combined with an explanation e into a sequence of tokens containing exactly one [MASK] token, as illustrated in Fig. 1(2). We report all patterns used in Appendix C.2.

One problem with generating an explanation $\hat{e}_{y'}$ for all $y' \in \mathcal{Y}$ is that explanations generated with false labels ($\hat{e}_{-y}$) are likely invalid. To allow the classification model to reason about these imperfect explanations, we fine-tune PET with explanations generated on all label conditions during training, and encourage the prediction to be the true label (y) regardless of the conditioning label. Our training objective minimizes the standard cross-entropy loss with explanation-aware patterns across all generated explanations:

$$\mathcal{L} = - \sum_{y' \in \mathcal{Y}} \log p_\theta \left( y \mid \text{EP}(x, \hat{e}_{y'}) \right),$$

with $p_\theta$ being the normalized probability from $f_\theta$.

---

[1] Labels can still appear in the prompt if they are positioned after explanations.

[2] We omit *explain-then-predict* from Fig. 1 for clarity. Conceptually, *explain-then-predict* is independent of the conditioning label, so the probability matrix in Fig. 1(2) would have identical rows and the rest of the pipeline is identical to *predict-then-explain*.

We choose this supervision objective because we hypothesize that it would be an effective way to leverage potentially unreliable explanations. For example, even degenerate explanations conditioned on wrong labels may suggest that GPT-3 have trouble justifying the incorrect label, thereby providing signals for the correct prediction. During inference, **FLamE** tries all generated explanations for a given instance, and makes the final prediction based on the label with the largest logit overall (Fig. 1(2)). Formally, we use the following prediction rule:

$$\hat{y} = \arg\max_{y \in \mathcal{Y}} \left( \max_{y' \in \mathcal{Y}} f_\theta \left( y \mid \text{EP}(x, \hat{e}_{y'}) \right) \right).$$

## 3 Experimental Setup

In this section, we present our experimental setup and discuss important choices in implementation. We will release our code upon publication.

### 3.1 Datasets

We need access to explanations in the test set to evaluate the quality of generated explanations in addition to task performance. We thus consider two natural language inference (NLI) tasks with natural language explanations:

- **e-SNLI** provides crowd-sourced free-form explanations for SNLI (Camburu et al., 2018).

- **e-HANS** offers templated explanations for HANS (Zhou and Tan, 2021). HANS is a templated NLI dataset designed to address syntactic heuristics in NLI tasks with 118 templates.

We focus on a few-shot learning setting with k=16 training examples and 16 development examples for each label class. We choose this moderate size (<100 examples for 3-class e-SNLI) because the number would be small enough to annotate for a new task, but also sizable for fine-tuning generation and classification models.

### 3.2 Baselines and Oracles

We use GPT-3 for explanation generation and choose RoBERTa (355M) as the underpinning prompt-based classifier (Brown et al., 2020; Liu et al., 2019b). To validate the effectiveness of **FLamE** against vanilla RoBERTa and PET, we include both methods without explanations as baselines. We further report classification performance of fine-tuned GPT-3 when explanations are not provided. We refer to these approaches as *no-explanation* as they do not use any explanations.

To demonstrate the inadequecy of the naive approach of using human explanations, namely, training with explanations and testing without, we report RoBERTa and PET results under this setting, referred to as *train-with-explanation*.

The explanation generation methods *explain-then-predict* and *predict-then-explain* also produce labels along with explanations, and are used in Wei et al. (2022b) and Lampinen et al. (2022). We thus include them as baselines. Recall that an important distinction in **FLamE** is that we use the generated explanations to fine-tune the prompt-based classification model so that it learns to leverage signals in unreliable explanations.

Finally, to examine the upper bound of classification with learning from explanations, we explore a condition in which we provide human explanations at inference time (*oracle-explanation*).

### 3.3 Implementation

We fine-tune two variants of GPT-3 models, Babbage and Davinci, as both explanation generators and classification baselines. We use vanilla (non-instruct) GPT-3 models, i.e., `babbage` and `davinci` in the API, because the InstructGPT variants are not available for fine-tuning. We use fine-tuned models for most results of the paper for two reasons. First, we find largely negative empirical results when generating explanations in-context using smaller models (e.g., GPT-3 Babbage). Second, for our choice of $k = 16$, fine-tuning is much cheaper than in-context learning.[3]

---
[3]Cost for GPT-3 APIs are calculated per-token. Fine-tuning eliminates the need for a prompting context and thus require significantly fewer tokens per inference.

Specifically, at training time, we fine-tune a GPT-3 model on $k \cdot |\mathcal{Y}|$ examples, with ground truth labels and human explanations encoded in the prompt. Refer to Appendix C.1 for GPT-3 generation prompts used in our experiments and hyperparameters used in fine-tuning GPT-3.

With the generated explanations, we fine-tune an explanation-aware prompt-based RoBERTa-large model under the PET framework. To ensure the premise and hypothesis are used by models, we ensemble **FLamE** with its *no-explanation* counterpart. We find that ensembling improves performance across the settings.

When tuning the classifier, we can choose to either incorporate gold explanations or explanations generated on the training set. We explore this choice as a hyperparameter, and find training with both generated explanations and gold explanations to be more effective than training exclusively on gold explanations for e-SNLI, and training with gold explanations is more effective for e-HANS. See Appendix C.3 for detailed results.

To contextualize our results, we list the number of parameters in models used in this work: GPT-3 Babbage (1.3B), GPT-3 Davinci (175B), and RoBERTa-Large (355M). As OpenAI does not publicly disclose GPT-3 parameters, we use estimates provided by Gao (2021).

## 4 Results

We demonstrate that our framework on learning from explanations is effective as it reliably outperforms baselines across datasets and conditions (4.1), and we analyze why and how explanations are useful in our framework (4.2, 4.3).

### 4.1 Classification Performance

Table 1 shows our main classification results. We start by comparing **FLamE** with the best performing baseline. Among the baselines, *no-explanation* achieves the best performance: GPT-3 Davinci achieves an accuracy of 78.6% in e-SNLI and PET has an accuracy of 70.7% in e-HANS. **FLamE** leads to a 5.7% improvement in e-SNLI as well as a 1.2% improvement in e-HANS, both achieved by *predict-then-explain* with explanations generated by GPT-3 Davinci.

|  |  | e-SNLI | | e-HANS | |
|---|---|---|---|---|---|
|  |  | Babbage | Davinci | Babbage | Davinci |
| *no-explanation* | RoBERTa (Liu et al., 2019b) | 49.4 | - | 57.5 | - |
|  | PET (Schick and Schütze, 2020) | 78.3 | - | *70.7* | - |
|  | GPT-3 (Brown et al., 2020) | 56.0 | *78.6* | 60.5 | 60.6 |
| *train-with-explanation* | RoBERTa | 39.5 | - | 47.5 | - |
|  | PET | 60.5 | - | 47.4 | - |
| *explain-then-predict* | GPT-3 (Wei et al., 2022b) | 33.6 | 50.6 | 63.6 | 57.6 |
|  | **FLamE** | 68.4 | 73.3 | 70.5 | 69.0 |
| *predict-then-explain* | GPT-3 (Lampinen et al., 2022) | 60.3 | 70.1 | 60.4 | 55.7 |
|  | **FLamE** | 77.9 | **84.3** | 64.1 | **71.9** |
| *oracle-explanation* | **FLamE** | 94.5 | - | 100.0 | - |

Table 1: Results on e-SNLI and e-HANS (k = 16). GPT-3 models are fine-tuned, so the implementation is slightly different from Wei et al. (2022b) and Lampinen et al. (2022). The column label Babbage and Davinci only apply to methods that use GPT-3, and is not relevant for RoBERTa and PET. Italicized numbers are from the strongest baselines and bolded are from the best **FLamE** set-up.

Next, we compare **FLamE** with two other approaches that learn from explanations to showcase its advantage. If we do not generate explanations, we do not have access to explanations at test time. Due to the distribution shift, we observe a large performance drop for PET *train-with-explanation*: the accuracy is 60.5% (e-SNLI) and 47.4% (e-HANS). RoBERTa *train-with-explanation* only provides an accuracy of 39.5% in e-SNLI. As a result, **FLamE** outperforms these approaches by more than 20%.

The more interesting comparison is with the counterpart that only uses GPT-3. For *explain-then-predict*, **FLamE** is always better than GPT-3, with improvements ranging from 6.9% to 34.8%. Similarly, for *predict-then-explain*, **FLamE** consistently outperforms GPT-3, with improvements ranging from 3.7% to 16.2%. In fact, GPT-3 *explain-then-predict* and *predict-then-explain* both result in performance drops from GPT-3 *no-explanation* in six out of eight cases. These results show that without prompt-based classification, GPT-3 cannot effectively use its own generated explanations, likely due to their unreliability.

Since users may not have access to the largest GPT-3 model due to financial considerations, we compare **FLamE** with both Babbage and Davinci. With Babbage, **FLamE** outperforms the second best approach by 17.6% in e-SNLI and 6.9% in e-HANS. With Davinci, **FLamE** outperforms the second best approach by 5.7% in e-SNLI and 11.3% in e-HANS. These improvements highlight the effectiveness of using a relatively small model to control a much bigger model (recall that RoBERTa-large has only 0.3% of parameters compared to Davinci).

Our result also shows that *predict-then-explain* generates more useful explanations than *explain-then-predict* prompts on e-SNLI as reflected in classification accuracy (+11.5% for Babbage, and +10.0% for Davinci) in Table 1. This result differs from Wei et al. (2022b)'s finding that post-answer explanations are not as effective as pre-answer explanations. The reason may be that natural language inference leads to different explanations from arithmetic reasoning. Explanations in Wei et al. (2022b) are procedural, and are more similar to instructions rather than explanations that provide proximal mechanisms (Tan, 2022). Thus, *explain-then-predict* may be more effective for such reasoning. In comparison, *predict-then-explain* leads to multiple different explanations generated for each example. Having access to multiple explanations at inference time increases the likelihood of having one that provides a strong signal for the true label.

We point out that supplying oracle explanations at both training and testing time leads to 94.5% on accuracy on e-SNLI and 100% accuracy on e-HANS. The numbers show that the information in explanations is helpful for classification if extracted effectively and there is room for further improvement by learning from explanations.

|  | Logical Consistency | Correct Template | Validity of Assumption |
|---|---|---|---|
| *predict-then-explain* | | | |
| e-SNLI (ê$_y$) | 45.0 | 95.0 | 58.3 |
| e-SNLI (ê$_{-y}$) | 15.0 | 75.0 | 71.7 |
| e-HANS (ê$_y$) | 42.0 | 76.9 | 75.2 |
| e-HANS (ê$_{-y}$) | 24.7 | 60.7 | 73.3 |
| *explain-then-predict* | | | |
| e-SNLI (ê) | 55.0 | 66.7 | 80.0 |
| e-HANS (ê) | 51.6 | 28.3 | 61.6 |

Table 2: Evaluation on explanations generated with GPT-3 Davinci (k = 16). ê$_y$ refer to explanations generated with ground-truth labels, and ê$_{-y}$ are explanations generated with false labels. For *explain-then-predict*, there is no conditioning label. See Table 6 in appendix for GPT-3 Babbage results.

In summary, for both PET and GPT-3 Davinci, learning from explanations hurts the performance compared to their *no-explanation* counterpart due to the absence of test-time explanations or/and the unreliable generation of explanations. **FLamE** addresses the unavailability of test-time explanations through generating explanations with GPT-3 and addresses the unreliable generation of explanations through prompt-based fine-tuning.

### 4.2 Explanation Evaluation

Ideally, the success of **FLamE** is driven by the successful generation of valid explanations. To understand why explanations are helpful for models, we first evaluate the quality of generated explanations with human evaluation. We formulate the following three criteria to evaluate both the content and the structure of generated explanations.

- Content-wise, *logical consistency* measures whether the explanation supports the true label with respect to the hypothesis given the premise.
- *Validity of assumption*, a relaxed version of logical consistency, measures whether the explanation shows understanding of the premise.[4]
- On the structure level, *correct template* measures whether the explanation includes matching label-specific cues (e.g., "not know" for neutral and "implies" for entailment) for the label that was used for generation. Table 3 shows an example

---
[4]If the generated explanation is irrelevant to the premise, then we consider it invalid.

| Premise | Supposedly the engineer expected the worker. |
|---|---|
| Hypothesis | The engineer expected the worker. |
| Label | Neutral |
| ê$_{ent}$ | Supposedly suggests the engineer expected the worker happened. |
| ê$_{neu}$ | Supposedly suggests an uncertainty, so we do not know whether the engineer expected the worker. |

Table 3: A label-specific cue for neutral examples is "not know" in the explanations, because the gold explanations for neutral examples always contain "not know." In this example, neutral-generated explanation contains this cue, whereas entailment-generated explanation does not. The classifier could predict neutral when "not know" is present in the generated explanation.

for label-specific cues. We use label-specific cues and templates interchangeably henceforth.

We annotated 20 generated examples (each with 3 explanations in e-SNLI and 2 explanations in e-HANS) for each test condition, with an inter-annotator agreement of 0.7 among three authors, measured by Krippendorff's alpha.

The quality of generated explanations is generally low. The majority of explanations are not logically sound, as logical consistency rarely surpasses 50% (Table 2). Validity of assumption scores reveal that explanations show understanding of premises most of the time, but they fail to connect premises and hypotheses correctly.

While the generated logic is bad, explanations show great promise in generating the correct label-specific cues. In fact, correct template scores are able to reach 95% and consistently exceed 60% with one exception. Therefore, template generation is likely associated with the performance improvement brought by **FLamE**. We include more analysis in Appendix B.

To sum up, generated explanations include invalid logic but can produce correct templates. These observations lead to our hypothesis that templates are driving classification, which we directly test in Section 4.3.

|  |  | $P(\hat{y}' \neq \hat{y}|e')$ | $P(\hat{y}' \neq \hat{y}|e'_1, e'_2, ...)$ | $P(y'_{\text{gen}} \neq y_{\text{gen}}|e'_1, e'_2, ...)$ |
|---|---|---|---|---|
| e-SNLI | Other item | - | 7.5 | 57.8 |
|  | N./V. replacement | 4.5 | 4.5 | 45.2 |
| e-HANS | Other item | - | 11.5 | 33.5 |
|  | N./V. replacement | 0 | 0 | 1.5 |

Table 4: Measures how often $\hat{y}$ (prediction) or $y_{\text{gen}}$ (label for generating the explanation that leads to the largest logit) changes given the modified explanations at inference time. We test on **FLamE** *predict-then-explain* models, and the original explanations are generated using GPT-3 Davinci.

### 4.3 Template-based Explanation Probe

To validate the role of label-specific cues, we modify explanations at test time and examine how much the changes affect predictions. In particular, we replace test-time explanations using:

- *Other-item explanations*: explanations generated for a different example with the same label.
- *Noun/verb replacement*: nouns and verbs of certain part-of-speech tags are randomly replaced in the explanation that leads to the largest logit.[5]

Both replacement methods preserve template information. *Other-item explanation* essentially shuffles test explanations among examples with the same label, so it preserves the template distribution over the entire test set as well as label-specific cues for the same label. However, it does not preserve templates used in each example since different templates may be used in explanations in different examples. *Noun/verb replacement*, more fine-grained, preserves templates for each example.[6]

How much replaced explanations change the prediction process shows the effect of label-specific cues on our model. Specifically, we measure the change in predicted label ($\hat{y}$) when we switch to a modified set of test explanations ($e'_1, e'_2, ...$) or make prediction only using the one altered explanation ($e'$) in the case of noun/verb replacement. Recall that each label is used to generate an explanation in *predict-then-explain*. Therefore, the set of modified explanations for noun/verb replacement explanations consist of one altered explanation and unaltered explanations. We also measure how often the largest logit comes from an explanation generated with a different label when we introduce the changes in test-time explanations. Finally, to account for randomness during replacement, we experiment with five seeds to replace explanations.

---

[5]We randomly replace tokens with one of the following part-of-speech tags: "NN","NNS","NNP", and "VBG".

[6]An example of this perturbation could be: "The man is smiling, not frowning" → "The sailor is creating, not working".

Surprisingly, these changes in test time explanations have little effects on predictions (Table 4). Testing on noun/verb-replaced explanation ($e'$) and discarding the unaltered explanations, we find that predictions do not change at all for e-HANS, and only changes 4.5% of the time for e-SNLI.

We find the effect on prediction small even if we test with all generated explanations for each example instead of using just $e'$. In fact, testing with noun/verb-replaced explanation does not change e-HANS predictions at all. The change in prediction is only 4.5% and 7.5% for the two replacement methods on e-SNLI, and it is only 11.5% for e-HANS other-item explanation.

While predicted labels do not vary much when explanations are perturbed, empirical evidence shows that the explanation used to generate the largest logit is conditioned on a different label for about half of the time on e-SNLI. In particular, for noun/verb replacement explanations, **FLamE** abstain from using the modified explanation 45.2% of the time. We think e-HANS does not have this property due to the templated nature of the dataset, which makes models more easily to pick up and even more heavily rely on the label-specific cue (i.e., "`not know`").

### 4.4 Where Does Classification Improvement Come From?

We find that classification improvement is two-fold: (1) GPT-3 generated explanations provide means for knowledge distillation; (2) Our RoBERTa-based classifier learns to distinguish which label is associated with the generated explanations.

In particular, our method is better than using GPT-3 alone to learn from explanations and predict labels (§4.1). This finding suggests that GPT-3 cannot effectively use its own generated explanations, likely due to the unreliability of generated explanations. Our probing experiments in §4.3 suggest that label-specific patterns are important, but we acknowledge that they may not be the only signal that the smaller model is able to extract.

|         | e-SNLI | e-HANS |
|---------|--------|--------|
| Babbage | 35.7   | 47.5   |
| Davinci | 76.2   | 85.7   |

Table 5: GPT-3 in-context learning results with k = 16.

If the label-specific cues drive the utility of explanations, one may wonder why we do not just identify those cues and use them instead of explanations. We argue that it is unclear what the cues can be (if the dataset is not constructed with templates, e.g., e-SNLI) when we only have few-shot explanations. Even in §4.3, where we did the template-based experiment, we treat everything except for nouns and verbs as "templates". On the other hand, our method learns from explanations and generates ones that provide useful cues for the downstream small classification model.

Overall, our framework provides a way to leverage information from LLMs, and we encourage future work to explore other possible approaches. For example, future work could examine ways to automatically extract useful signals from LLM-generated auxiliary inputs.

## 5 GPT-3 In-Context Learning

Since OpenAI reduced its API pricing, the authors decided to obtain in-context learning results for GPT-3 *no-explanation*. Table 5 shows that GPT-3 Babbage in-context learning does not perform well on the datasets, and **FLamE** (with Babbage generated explanations) easily outperforms it by a huge amount (+42.2% on e-SNLI and 31.8% on e-HANS).[7] This observation is consistent with our preliminary experiments that suggest fine-tuning outperforms in-context learning on Babbage.

Even if we increase GPT-3 model size to 175B (Davinci), **FLamE** still outperforms in-context learning on e-SNLI (+8.1%). Similar to Babbage, fine-tuning provides better performance than in-context learning in e-SNLI. In contrast, GPT-3 Davinci in-context learning performs better on e-HANS, likely due to its templated nature. According to the induction heads hypothesis (Olsson et al., 2022), in-context learning uses two kind of attention heads to copy and complete patterns. GPT-3 Davinci may utilize this mechanism to achieve high performance on e-HANS.

---
[7]In-context learning experiments are done with the Instruct-GPT (Ouyang et al., 2022) series, namely `text-babbage-001` and `text-davinci-002`.

The divergent behavior between fine-tuning and in-context learning requires additional investigation. It further motivates research on controlling these black-box models that are not easily accessible to the majority of researchers.

## 6 Related Work

We review additional related work in natural language explanations (NLEs), few-shot learning, and model distillation.

**Generating and using natural language explanations.** A variety of previous studies examine the generation of NLEs via fine-tuning generative language models or prompting LLMs (Narang et al., 2020; Nye et al., 2021; Marasović et al., 2022; Wang et al., 2022b). A natural way of using NLEs is to build models with explanations in order to increase performance or robustness (Hancock et al., 2018; Rajani et al., 2019; Zhou and Tan, 2021; Mishra et al., 2022).

With the advent of LLMs, additional approaches for learning from NLEs emerge. Wei et al. (2022b) incorporate step-by-step NLEs into a *chain-of-thought* prompt and demonstrate its effectiveness on certain benchmarks. Zelikman et al. (2022) use LLMs to generate rationales and further finetune LLMs on the generated explanations to improve performance over LLMs trained without rationale. Meanwhile, Lampinen et al. (2022) observe limited gains by adding NLEs post-answer to in-context learning. Our approach is different in that we use LLMs to generate explanations rather than making predictions, and train a separate model to overcome the unreliability of generated explanations.

The strong abilities of LLMs also lead to a lot of recent work on leveraging them to generate part of the input for a separate model. Ye and Durrett (2022) evaluate the factuality of GPT-3 generated explanations and calibrate models with factuality scores. Our framework does not require additional explanation evaluation scores for calibration and achieves higher accuracy improvement. In addition, Meng et al. (2022) use GPT-2 to generate class-conditioned *hypotheses* given premise and labels as training data for RoBERTa. In comparison, our framework learns from *explanations* by using GPT-3 to generate explanations and a smaller model for label prediction. We preserve the original NLI input and conduct in-depth analysis to understand the performance improvement.

Moreover, LLMs have been leveraged to generate intermediate context for commonsense reasoning and question answering. Some work (Liu et al., 2022a; Wang et al., 2022a) uses LLM outputs to train a smaller model that generates knowledge. Paranjape et al. (2021) prompt LLMs to generate contrastive explanations to improve performance. In a similar vein, Liu et al. (2022b) uses LLM to generate knowledge for commonsense reasoning tasks. External knowledge can be crucial for commonsense reasoning, so these works focus on generating knowledge to improve performance, whereas our work focus on generating explanations for inference tasks.

An additional motivation for using NLEs is to improve the explainability of in-context learning. Min et al. (2022) show that in-context learning classification performance drops only marginally after replacing gold labels in the demonstrations to random labels. Generating explanations for the labels provides additional information for classification, whether being used as reasoning (e.g., chain-of-thought) or as input to a calibrator (e.g., our approach). Note that we do not imply that such explanations are faithful to the actual computation in the model (Turpin et al., 2023).

NLEs also have broad applications beyond language, such as visual reasoning, reinforcement learning, and solving algebraic word problems (Hendricks et al., 2016; Park et al., 2018; Zellers et al., 2019; Hernandez et al., 2022; Ling et al., 2017; Andreas et al., 2017).

**Few-shot learning.** Underlying our explanation-aware classifier, Pattern-Exploiting Training (PET) (Schick and Schütze, 2020) converts few-shot classification to mask infilling. Similarly, Gao et al. (2020) incorporates demonstration examples into prompt-based fine-tuning. A related line of work treats LMs as knowledge bases (Trinh and Le, 2019; Petroni et al., 2019). Under this framing, few-shot learning boils down to identifying good queries, which often come in the form of carefully constructed prompts (Radford et al., 2019; Jiang et al., 2020; Brown et al., 2020; Le Scao and Rush, 2021). Earlier work on few-shot learning applies techniques in semi-supervised training such as data augmentation (Miyato et al., 2017; Clark et al., 2018; Xie et al., 2020a). Our work provides a few-shot learning framework for learning from explanations by combining LLMs and prompt-based classification.

**Model Distillation.** The training of a separate RoBERTa-based model can also be interpreted as model distillation through NLEs. There has been a lot of work on distilling knowledge in neural networks (Hinton et al., 2015; Liu et al., 2019a; Xie et al., 2020b). The most related work is in context distillation (Snell et al., 2022; Choi et al., 2022; Askell et al., 2021), where models are trained to internalize step-by-step reasoning, but they do not address the absence of high-quality reasoning during test time.

## 7 Conclusion

We present **FLamE**, a two-stage framework that leverages the few-shot generation capability of GPT-3 and a relatively small model to effectively use the generated explanations with fallible reasoning. Our approach outperforms strong baselines in natural language inference. We further show that while the generated explanations are invalid, they include useful label-specific cues. Through a probing experiment, we prove that these label-specific cues are essential for model prediction.

We believe that using a smaller model to leverage the outputs from large language models is a promising direction for future work. This approach has at least two advantages: 1) the small model can potentially handle the imperfect outputs from the large model; 2) the small model allows for efficient interpretation and probing of the final pipeline. Future work may investigate removing the dependency on the large model altogether at test time.

## Limitations

Our work focuses on building a two-stage framework for generating and learning from explanations. In our investigation, we are limited by the available computational resources, financial budgets, and datasets. GPT-3 and PET are performant few-shot learners that work well for our use case. However, GPT-3 is not free to use and partly for financial considerations, we did not experiment with GPT-3 in-context learning initially. The performance difference between GPT-3 Babbage and Davinci are aligned with the emergent abilities of large-scale language models (Wei et al., 2022a; Rae et al., 2022). Therefore, in the era of research with private large-scale language models, it would be useful for the research community to collectively build knowledge about how large-scale language models work. It would be useful to experiment with other models such as Google's PaLM (540B) (Chowdhery et al., 2022b) and Deepmind's Gopher (280B) (Rae et al., 2022). It is an important question for the research community to explore productive paths forward.

Often, prompt engineering requires either significant manual work to come up with good templates (Brown et al., 2020; Schick and Schütze, 2020) or a big budget to run automatic prompt generation methods (Lester et al., 2021; Wu et al., 2022). In this work, we used a fixed prompt (see Appendix C.1) for explanation generation, future work could also investigate from the angle of generating better prompts.

We experimented with two natural language inference tasks, which tend to correlate with a certain form of explanations. One way to interpret the difference in our findings and chain-of-thought prompting is indeed that the reasoning in e-SNLI and e-HANS are not the multi-step reasoning used in arithmetic reasoning. As Tan (2022) argues, there are diverse types of explanations, which may lead to varying levels of effectiveness from a learning method. Future work could investigate the effectiveness of our method on other tasks and different types of explanations.

While our method demonstrates effectiveness against strong baselines, there is still a big gap from the upper bound performance and suggests potential for better use of the explanations in future work. For example, future work could incorporate careful example selection into learning with explanations. We picked examples randomly, but research has shown that calibration (Zhao et al., 2021) reordering (Lu et al., 2022) and example selection (Liu et al., 2021) changes GPT-3's behavior. We also used human explanations to fine-tune the GPT-3 model for explanation generation, but human explanations may not always be high-quality or the best guide for machine learning models.

Additionally, we use RoBERTa as our backbone model for the classifier used in both the non-GPT baselines and our **FLamE** framework. We manage to beat strong GPT-3 baselines that use explanations. While more powerful classifiers (e.g., DeBERTa) could also be used in place of RoBERTa, we believe we have demonstrated the effectiveness of our method by using a simpler classifier. We leave it to future work to investigate the effectiveness of our method with more powerful classifiers.

Finally, it is worth noting that we use a particular setup of k = 16 for our experiments. While we believe that this is a reasonable few-shot learning setup, results could differ for different k. We leave it to future work for examining the impact of examples, explanations, and number of samples.

## Broader Impacts

We propose a framework to generate and learn from explanations and conduct in-depth analysis to understand the utility of explanations. Our work has the potential to help people understand the behavior or usage of large-scale language models and improve their trustworthiness.

## Acknowledgements

We thank Sherry Tongshuang Wu and the members of the Chicago Human+AI Lab for their insightful feedback. We also thank anonymous reviewers for their helpful suggestions and comments. This work is supported in part by an NSF grant, IIS-2126602.

## A  An Overview of Pattern-Exploiting Training (Schick and Schütze, 2020)

The essence of PET is to reduce classification to mask infilling. A pre-defined *pattern* $P : \mathcal{X} \to \mathcal{V}^\star$ converts a task instance $x$ into a sequence of tokens $P(x)$ in the vocabulary $\mathcal{V}$, under the restriction that $P(x)$ contains exactly one masked token. PET further utilizes a *verbalizer* $V$, which declares a special set of tokens, each representing a label in the label set. Then, classification, choosing one label from the label set, boils down to infilling one token in this special set. Formally, the *verbalizer* $V : \mathcal{Y} \to \mathcal{V}$ is an injective map from the label set $\mathcal{Y}$ to the model's vocabulary $\mathcal{V}$.

With these tools defined, PET is formulated as

$$\hat{y} = \underset{y \in \mathcal{Y}}{\arg\max}\, f_\theta\left(V(y) \,|\, P(x)\right),$$

where $f_\theta(t|s)$ is the (unnormalized) probability of unmasking token $t$ from the sequence $s$ which contains exactly one masked position. For simplicity, our formulation only assumes one *pattern-verbalizer* pair (PVP), and uses the unweighted average of logits from multiple PVPs in implementation. We further simplify PET by removing the distillation and the multi-task learning objective, as we find these extensions have marginal impacts on performance but are costly in computation.

|  | Logical Consistency | Correct Template | Validity of Assumption |
|---|---|---|---|
| *predict-then-explain* | | | |
| e-SNLI ($\hat{e}_y$) | 28.3 | 73.3 | 61.7 |
| e-SNLI ($\hat{e}_{-y}$) | 3.3 | 70.8 | 52.5 |
| e-HANS ($\hat{e}_y$) | 54.6 | 71.9 | 87.4 |
| e-HANS ($\hat{e}_{-y}$) | 27.6 | 64.8 | 82.6 |
| *explain-then-predict* | | | |
| e-SNLI ($\hat{e}$) | 11.7 | 16.7 | 63.3 |
| e-HANS ($\hat{e}$) | 59.9 | 69.5 | 84.5 |

Table 6: Evaluation on explanations generated with GPT-3 Babbage ($k = 16$). $\hat{e}_y$ gives evaluation on explanations generated with ground-truth labels, and $\hat{e}_{-y}$ gives evaluation on explanations generated with false labels. For *explain-then-predict*, the generated explanation is not conditioned on any label.

## B  Error Analysis

Although explanations are logically incorrect most of the time, the classification model manages to take them as inputs and correctly predict the label. To understand why illogical explanations are useful, we conduct an error analysis by comparing PET *no-explanation* baseline and **FLamE** (*predict-then-explain*) errors. We generate the confusion matrix over the test set and measure properties of explanations in each component (Table 13).

In both e-SNLI and e-HANS, the confusion matrix is heavy along the diagonals, suggesting that **FLamE** and PET *no-explanation* agree most of the time. Breaking down the improvement by class, **FLamE** improves e-SNLI mostly in the contradiction (42.9%) and neutral (45.1%) examples. Whereas e-HANS improvements mostly come from the entailment class (53.8%).

To examine the explanations, we use BLEU scores[8] to measure similarity beween generated explanations and ground truth. In addition, for e-HANS, where ground-truth explanations always contain "`not know`" for the "neutral" class, we compute the rate of correctly generating "`not know`" to measure template similarity between generated explanations and ground truth.

We find that **FLamE** is more likely to make correct predictions when the generated explanations are similar to ground truth in e-HANS as illustrated by the BLEU scores in Table 13. Our qualitative analysis on 5 examples sampled from e-HANS errors confirms this finding (Table 14,15).

Not only are generated contents similar to the ground-truth explanations when **FLamE** makes correct predictions, generated *templates* are also similar to ground truth. In fact, examples in (**FLamE** ✓, *no-explanation* ✗) perfectly and accurately generate "`not know`", whereas examples in (**FLamE** ✗, *no-explanation* ✓) only correctly generate "`not know`" 15% of the time. This finding suggests that prediction accuracy is correlated with the correctness of generating "`not know`" and further motivates our analysis at how much templates can affect our model.

We also measure *label consistency*, that is,

---
[8]We use uniform weights and compute BLEU-4. Since explanations are usually short in length, we use a smoothing function (Chen and Cherry, 2014).

| | |
|---|---|
| Premise | if the essayist smiled , the photojournalist avoided the programmer . |
| Hypothesis | the essayist smiled . |
| Label | neutral |
| ê$_{ent}$ | the photojournalist avoided the programmer if the essayist smiled , we do not know whether the essayist smiled . |
| ê$_{neu}$ | if the essayist smiled , the photojournalist avoided the programmer . |

Table 7: e-HANS example where label consistency is not met. **FLamE** uses ê$_{ent}$ to predict the correct label "neutral".

whether the predicted label is the same as the label used to generate the explanation that leads to the largest logit. High label consistency means explanations generated with the predicted label also gives the best utility in predicting that label. It also shows whether GPT-3 is able to generate useful explanations given the correct label.

We find that **FLamE** uses the explanations generated with the predicted label most of the time for both e-SNLI (>65%) and e-HANS (>70%). However, there are still instances where GPT-3 generates better explanations with a wrong label (Table 7). In particular, only 38.5% of e-HANS examples in the (**FLamE** ✓, *no-explanation* ✗) category achieves label consistency.

## C  Implementation Details

### C.1  GPT-3 Prompts & Hyperparameters

Following (Wiegreffe et al., 2022), we adopt a minimalistic prompt design for e-SNLI and e-HANS. We report prompts for both datasets in Table 10. GPT-3 fine-tuning hyperparameters are shown in Table 8. We followed recommended hyperparameters by OpenAI and they worked well by eyeballing.

### C.2  PET PVPs & Hyperparameters

We append explanations to existing PET patterns and show our explanation-aware pattern verbalizer pairs in Table 11. PET hyperparameters are shown

| Hyperparameter | |
|---|---|
| Train Epochs | 10 |
| Batch Size | 4 |
| Learning Rate Multiplier | 0.1 |

Table 8: List of hyperparameters used when fine-tuning GPT-3.

| Hyperparameter | |
|---|---|
| Train Steps | 1000 |
| Batch Size | 4 |
| Beta initial value | {0.0, 0.25, 0.5, 0.75, 1.0} |
| Beta learning rate | {2e-2, 2e-3, 2e-4} |
| Training explanation | {generated expl., ground-truth expl., gold-label generated ($\hat{e}_y$), generated ∪ ground-truth, $\hat{e}_y$ ∪ ground-truth} |

Table 9: List of hyperparameters used when fine-tuning PET.

in Table 9.

### C.3 Training with different explanations

We show **FLamE** results on e-SNLI and e-HANS when trained with different set of explanations in Table 12.

### C.4 GPU Decision

For all experiments reported in the paper, we use A40. In preliminary experiments, we find that RTX8000 and A40 can produce different results. So for replicability, one should run our code on A40s.

## D Human Evaluation on GPT-3 Babbage Explanations

See evaluation results in Table 6. Similar to GPT-3 Davinci generated explanations, these explanations are largely illogical in supporting the ground-truth label but show understanding of the premise relatively well. In addition, these explanations can mostly correctly generate label-specific cues, except for explanations generated for e-SNLI with *explain-then-predict* prompts.

| Dataset | Prompt |
|---------|--------|
| e-SNLI | ```
Three people on a ski trail on a sunny day.
question: There is nine feet of snow on the ground.
maybe
why?
###
Not all ski trail has nine feet of snow on the ground.
###
``` |
| e-HANS | ```
the manager that helped the technician addressed the illustrator .
question: the manager helped the technician .
true
why?
###
that in that helped the technician refers to the manager .
###
``` |

Table 10: Examples of prompts for e-SNLI and e-HANS. During fine-tuning, GPT-3 models are given the `premise`, `hypothesis` and a `conditioning label` in the prompt, while the `ground truth explanation` is used as the generation target. During inference, we still provide the `premise`, `hypothesis` and a `conditioning label`, while eliciting a `generated explanation` from the fine-tuned model. We include ``###'' in the prompt as explicit signals for explanation generation.

| Dataset | Verbalizer | Pattern |
|---------|------------|---------|
| e-SNLI | {yes, no, maybe} | "premise"?[mask], "hypothesis" because "expl" |
|         | {yes, no, maybe} | premise?[mask],hypothesis because expl |
|         | {right, wrong, maybe} | "premise"?[mask], "hypothesis" because "expl" |
|         | {right, wrong, maybe} | premise?[mask],hypothesis because expl |
| e-HANS | {yes, maybe} | "premise"?[mask], "hypothesis" because "expl" |
|         | {yes, maybe} | premise?[mask],hypothesis because expl |
|         | {right, maybe} | "premise"?[mask], "hypothesis" because "expl" |
|         | {right, maybe} | premise?[mask],hypothesis because expl |

Table 11: Explanation-aware pattern-verbalizer pairs.

|  | e-SNLI | | e-HANS | |
| --- | --- | --- | --- | --- |
|  | Babbage | Davinci | Babbage | Davinci |
| gen | | | | |
| **FLamE** *explain-then-predict* | 0.684 | 0.701 | 0.637 | 0.683 |
| **FLamE** *predict-then-explain* | 0.779 | **0.834** | 0.641 | 0.674 |
| gold | | | | |
| **FLamE** *explain-then-predict* | 0.671 | 0.709 | 0.705 | 0.69 |
| **FLamE** *predict-then-explain* | 0.755 | 0.782 | 0.637 | **0.719** |
| gold+gen | | | | |
| **FLamE** *explain-then-predict* | 0.669 | 0.729 | 0.7 | 0.686 |
| **FLamE** *predict-then-explain* | 0.761 | **0.843** | 0.641 | 0.657 |
| gold-gen | | | | |
| **FLamE** *explain-then-predict* | 0.66 | 0.71 | 0.705 | 0.69 |
| **FLamE** *predict-then-explain* | 0.755 | 0.782 | 0.638 | **0.719** |
| gold+gold-gen | | | | |
| **FLamE** *explain-then-predict* | 0.669 | 0.733 | 0.637 | 0.683 |
| **FLamE** *predict-then-explain* | 0.757 | 0.782 | 0.641 | **0.718** |
| overall | | | | |
| **FLamE** *explain-then-predict* | 0.684 | 0.733 | 0.705 | 0.69 |
| **FLamE** *predict-then-explain* | 0.779 | **0.843** | 0.641 | **0.719** |

Table 12: **FLamE** results with different training explanations.

|  | e-SNLI | | | e-HANS | | | |
| --- | --- | --- | --- | --- | --- | --- | --- |
|  | % | BLEU | Label Consistency | % | BLEU | ``not know'' Correctness | Label Consistency |
| both ✓ | 75.2 | 9.7 \| 7.7 | 64.4 | 66.7 | 56.8 \| 40.6 | 88.6 | 74.5 |
| **FLamE** ✓, *no-explanation* ✗ | 9.1 | 8.9 \| 7.2 | 64.8 | 5.2 | 63.2 \| 58.5 | 100.0 | 38.5 |
| **FLamE** ✗, *no-explanation* ✓ | 3.1 | 10.4 \| 7.1 | 71.0 | 4.0 | 21.6 \| 18.6 | 15.0 | 75.0 |
| both ✗ | 12.6 | 10.1 \| 8.3 | 67.5 | 24.1 | 39.6 \| 21.4 | 66.4 | 83.8 |

Table 13: Error analysis comparing **FLamE** *predict-then-explain* with PET *no-explanation* baseline. BLEU scores take the format of (BLEU scores for the true label | BLEU scores for the false label).

| Examples |
|---|
| label:entailment<br>premise:before the writer advised the manager, the stylist encouraged the essayist.<br>hypothesis:the writer advised the manager.<br>human expl.:before suggests the writer advised the manager happened.<br>entailment gen. expl.:before suggests the writer advised the manager happened.<br>neutral gen. expl.:before suggests a swap happened. |
| label:entailment<br>premise:after the illustrator continued, the programmer advised the managers.<br>hypothesis:the programmer advised the managers.<br>human expl.:the illustrator continued, so the programmer advised the managers.<br>entailment gen. expl.:the illustrator continued and the programmer advised the managers, so the programmer advised the managers happened.<br>neutral gen. expl.:the illustrator continued and the programmer advised the managers, we do not know whether the programmer advised the managers. |
| label:neutral<br>premise:if the musician lay, the programmer performed.<br>hypothesis:the programmer performed.<br>human expl.:we do not know if the musician lay,<br>so we do not know if the programmer performed.<br>entailment gen. expl.:the musician lay if the programmer performed, we do not know whether the programmer performed.<br>neutral gen. expl.:the musician lay if the programmer performed, we do not know whether the programmer performed. |
| label:neutral<br>premise:unless the illustrator listened, the programmer existed.<br>hypothesis:the programmer existed.<br>human expl.:we do not know if the illustrator listened,<br>so we do not know if the programmer existed.<br>entailment gen. expl.:the illustrator listened if the programmer existed, we do not know whether the programmer existed.<br>neutral gen. expl.:we do not know if the illustrator listened,<br>so we do not know if the programmer existed. |
| label:entailment<br>premise:since the artisans smiled, the programmers continued.<br>hypothesis:the artisans smiled.<br>human expl.:since suggests the artisans smiled happened.<br>entailment gen. expl.:since suggests the artisans smiled happened.<br>neutral gen. expl.:since suggests the artisans smiled happened. |

Table 14: Test examples that **FLamE** gets correct but PET *no-explanation* does not.

| Examples |
|---|
| premise:the baker brought the colorist that the strategist advised.<br>hypothesis:the strategist brought the baker.<br>human expl.:the colorist is the object of the relative clause that the strategist advised and the baker brought the colorist, we do not know whether the strategist brought the baker.<br>entailment gen. expl.:that suggests the strategist brought the baker.<br>neutral gen. expl.:the strategist advised the colorist, so the strategist brought the baker. |
| premise:the artisans thanked the writer and the worker.<br>hypothesis:the writer thanked the worker.<br>human expl.:the artisans thanked the writer and the artisans thanked the worker, we do not know whether the writer thanked the worker.<br>entailment gen. expl.:the artisans are part of the worker.<br>neutral gen. expl.:the artisans thanked the writer and the worker, but the worker thanked the writer. |
| premise:whether or not the analyst encouraged the colorists, the bakers expected the planner.<br>hypothesis:the analyst encouraged the colorists.<br>human expl.:whether or not suggests we do not know whether the analyst encouraged the colorists.<br>entailment gen. expl.:the bakers expected the planner despite the clause whether or not the analyst encouraged the colorists.<br>neutral gen. expl.:the bakers expected the planner despite the clause whether or not the analyst encouraged the colorists. |
| premise:whether or not the officer expected the worker, the illustrators avoided the technician.<br>hypothesis:the officer expected the worker.<br>human expl.:whether or not suggests we do not know whether the officer expected the worker.<br>entailment gen. expl.:the illustrators avoided the technician despite the clause whether or not the officer expected the worker.<br>neutral gen. expl.:the illustrators avoided the technician despite the officer expected the worker. |
| premise:the officers by the psychiatrist saw the analyst.<br>hypothesis:the psychiatrist saw the analyst.<br>human expl.:the officers are by the psychiatrist and the officers saw the analyst, we do not know whether the psychiatrist saw the analyst.<br>entailment gen. expl.:the officers by the psychiatrist suggests the psychiatrist saw the analyst happened.<br>neutral gen. expl.:the officers by the psychiatrist saw the analyst, if the officers by the psychiatrist saw the analyst, then we do not know whether the psychiatrist saw the analyst. |

Table 15: Test examples that **FLamE** gets wrong but PET *no-explanation* gets correct. All the examples are from the neutral class.